\documentclass[10pt,twocolumn,letterpaper]{article}

 \usepackage[pagenumbers]{cvpr} 
\usepackage{booktabs}   
\usepackage{arydshln}   
\usepackage{multirow}
\usepackage{amsmath, amssymb, amsfonts}
\usepackage{threeparttable}  
\usepackage[english]{babel}
\usepackage{enumitem}

\definecolor{cvprblue}{rgb}{0.21,0.49,0.74}
\usepackage[pagebackref,breaklinks,colorlinks,allcolors=cvprblue]{hyperref}

\def\paperID{****} 
\def\confName{CVPR}
\def\confYear{2026}

\title{BrainStack: Neuro-MoE with Functionally Guided Expert Routing for EEG-Based Language Decoding}

\author{
    Ziyi Zhao$^1$, 
    Jinzhao Zhou$^1$, 
    Xiaowei Jiang$^1$, 
    Beining Cao$^1$, 
    Wenhao Ma$^1$, 
    Yang Shen$^1$, \\
    Ren Li$^2$, 
    Yu-kai Wang$^1$, 
    and Chin-Teng Lin$^{1,*}$ \\
    \noalign{\vskip 1ex} %
    \normalsize $^1$University of Technology Sydney, Sydney, Australia \\
    \normalsize $^2$Mohamed bin Zayed University of Artificial Intelligence, Masdar City, UAE \\
	\normalsize \texttt{ziyi.zhao-2@student.uts.edu.au, chin-teng.lin@uts.edu.au}
}

\begin{document}
\maketitle
\begin{abstract}
Decoding linguistic information from electroencephalography (EEG) remains challenging due to the brain’s distributed and nonlinear organization. We present BrainStack, a functionally guided neuro–mixture-of-experts (Neuro-MoE) framework that models the brain’s modular functional architecture through anatomically partitioned expert networks. Each functional region is represented by a specialized expert that learns localized neural dynamics, while a transformer-based global expert captures cross-regional dependencies. A learnable routing gate adaptively aggregates these heterogeneous experts, enabling context-dependent expert coordination and selective fusion. To promote coherent representation across the hierarchy, we introduce cross-regional distillation, where the global expert provides top-down regularization to the regional experts. We further release SilentSpeech-EEG (SS-EEG), a large-scale benchmark comprising over 120 hours of EEG recordings from 12 subjects performing 24 silent words—the largest dataset of its kind. Experiments demonstrate that BrainStack consistently outperforms state-of-the-art models, achieving superior accuracy and generalization across subjects. Our results establish BrainStack as a functionally modular, neuro-inspired MoE paradigm that unifies neuroscientific priors with adaptive expert routing, paving the way for scalable and interpretable brain-language decoding.
\end{abstract}

\section{Introduction}
\label{sec:intro}
\begin{figure*}[htbp]
    \centering
    \includegraphics[width=0.9\linewidth]{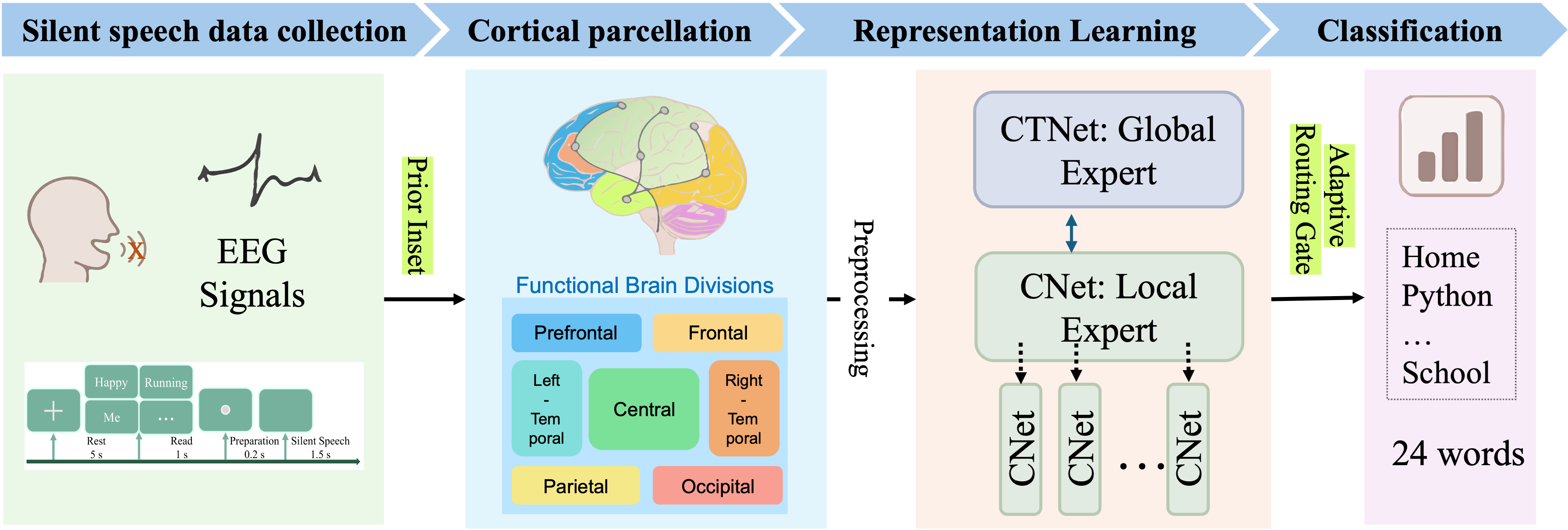}
    \caption{Overall illustration of EEG-based text decoding using BrainStack.}
    \vspace{-0.5em}  
    \label{fig:heatmap}
\end{figure*}

The human brain is a highly organized and distributed system composed of approximately 86 billion neurons. These neurons interact through complex electrical activity, and their collective dynamics give rise to macroscopic field potentials known as brainwave signals. Electroencephalography (EEG) provides a non-invasive and high-temporal-resolution technique for capturing large-scale cortical dynamics, making it particularly well-suited for real-time neural decoding \cite{he2013brain}. With millisecond-level precision, modern EEG acquisition systems enable scalable and practical recording of neural signals across distributed brain regions \cite{ lou2024dbnet}. These capabilities form the foundation for brain-computer interface (BCI) applications, especially in cognitive and communication tasks where timely and accurate decoding of brain activity is essential \cite{liang2024}.

Despite its potential, decoding meaningful information from EEG remains fundamentally challenging due to the low signal-to-noise ratio, temporal non-stationarity, and the inherent heterogeneity of neural activity across individuals and brain regions \cite{choi2025geometric, duan2023}.

Recent advances in deep learning have significantly improved the modeling capacity of EEG decoders. Convolutional and temporal models, such as CNNs, RNNs, and Transformers—have shown great potential in capturing complex spatiotemporal dependencies in EEG recordings \cite{wang2024cbramod, shi2023meet, ding2025attentive, duan2023replay, cao2022building}. Moreover, trends like large-scale pretraining and self-supervised learning, which revolutionized natural language and vision domains, are starting to demonstrate encouraging results in neural decoding\cite{jiang2024neurolm, wang2024eegpt, zhu2023eeg2vec, tang2024triplet, svanera2021selfsupervised, wu2022lowlight}. Despite this progress, most existing EEG models implicitly assume a homogeneous cortical structure, processing the entire electrode montage as a single unified source. This assumption overlooks the brain’s well-established functional modularity, where distinct cortical regions exhibit specialized dynamics and contribute differently to cognitive processes. As a consequence, global models often overfit to noise-rich whole-brain signals, while region-aware variants underutilize cross-regional complementarities \cite{liang2024transfer, chen2024robust, li2020unsupervised, deng2023seizure, hao2022speed}. To address these limitations, recent efforts have explored multi-band modeling, region-specific attention, subject-adaptive learning, and dynamic neural representations \cite{shevchenko2024comparative, bian2024ondevice, lou2024dbnet, duggento2022intertwined}. However, these approaches remain fundamentally monolithic—they lack an explicit mechanism to model the brain as a collection of interacting functional modules or to coordinate region-specific learners in a principled manner.

The latest methods have adopted channel-wise tokenization strategies—treating each electrode as an independent sequence unit fed into Transformer backbones \cite{lee2021eegtransformer, ding2025attentive}. While this formulation aligns with attention mechanisms and facilitates efficient temporal modeling, it inherently ignores anatomical clustering and spatial continuity. Consequently, it overlooks the physiologically grounded inter-regional dependencies that shape neural representations \cite{shi2023meet, choi2025geometric, zheng2024discrete}. This limitation becomes especially pronounced in silent-speech decoding, where fine-grained, cross-regional activation patterns are essential for distinguishing subtle internal articulatory states \cite{wang2020silent, rekrut2021semantic, song2023sEMG, fitriah2022survey}.

In contrast, neuroscience offers strong motivation to respect the brain’s functional heterogeneity. Neuroscientific research has long emphasized the modular organization of the brain, with distinct cortical areas associated with specific cognitive roles \cite{lashley2006organization}, especially language \cite{luoStableDecodingSpeech2023,jamaliSemanticEncodingLanguage2024,liuDecodingSynthesizingTonal2023}. Classical anatomical studies consistently parcellate the cortex into functional modules—prefrontal, temporal, parietal, and occipital regions—each associated with distinct cognitive pathways \cite{stephan2006dcm}.  Neural activity during cognitive and linguistic tasks is therefore spatiotemporal and distributed, with region-specific activation patterns reflecting executive control, articulatory planning, auditory–semantic processing, and visual integration \cite{jiang2025neural, anumanchipalliSpeechSynthesisNeural2019, cao2021brain, li2022unconsciousness}. These insights highlight the importance of functionally informed architectures that explicitly respect regional specialization to exploit the diversity of cortical dynamics and enhance generalization \cite{johnston2024modular}.
 While some recent studies have begun to incorporate region-aware modeling into EEG decoding, many still rely on simplified channel selection schemes or majority-voting ensembles, which are insufficient to capture the rich, cooperative dynamics across regions \cite{abbasi2020modularity, ding2022sensory}. Simple decision-level fusion strategies also struggle to generalize in low-data or high-noise regimes, where modeling the interaction between global and local representations becomes particularly important \cite{yue2017complexity}. Graph-based approaches such as LGGNet have made initial progress toward incorporating region priors, but they generally adopt homogeneous encoders and static connectivity patterns, limiting their ability to capture dynamic coordination between scales \cite{ding2023lggnet}. 

To overcome these limitations, we introduce BrainStack, a functionally guided Neuro–Mixture-of-Experts (Neuro-MoE) framework designed for robust EEG-based language decoding. 
Motivated by the brain’s modular organization, BrainStack partitions the EEG input into anatomically defined cortical regions and assigns each region to a specialized expert network, while a transformer-based global expert models long-range dependencies and whole-brain contextual interactions. Each regional expert uses a lightweight convolutional encoder to capture localized neural dynamics, whereas the global expert integrates broader temporal–spatial structure through a hybrid convolution–attention design. To coordinate these heterogeneous experts, BrainStack employs an adaptive expert routing gate that assigns context-dependent routing weights to expert representations and aggregates them through a learnable routing mechanism. 
To further enhance intra-expert coherence, we employ cross-regional hierarchical distillation in which the global expert provides top-down semantic guidance to regional experts. By aligning their output distributions, the distillation encourages each expert to learn not only region-specific cues but also the global semantic structure underlying silent-speech representations.

Our overarching objective is to develop an efficient and generalizable EEG decoder capable of exploiting complementary neural signals across distributed cortical modules. BrainStack achieves this by combining functional modularity, adaptive expert routing, and hierarchical distillation within a unified Neuro-MoE formulation—addressing limitations of prior homogeneous or weakly coupled EEG models and improving accuracy, scalability, and interpretability for practical BCI applications. To summarize, our contributions are as follows.

\begin{itemize}
\item We propose BrainStack, a neuroscientifically grounded Neuro-MoE architecture that integrates global and region-specific experts through anatomically guided modularization and adaptive expert routing.
\item We introduce a hierarchical cross-regional distillation mechanism in which the global expert supervises regional experts, promoting top-down semantic alignment and robustness under subject variability.
\item We present SilentSpeech-EEG (SS-EEG), a large-scale benchmark containing over 120 hours of EEG data from 12 subjects across 24 silent words—the largest dataset for word-level neural decoding to date.
\end{itemize}

\section{Methodology}
\label{sec:methodology}

\begin{figure}[htbp]
    \centering
    \includegraphics[width=1\linewidth]{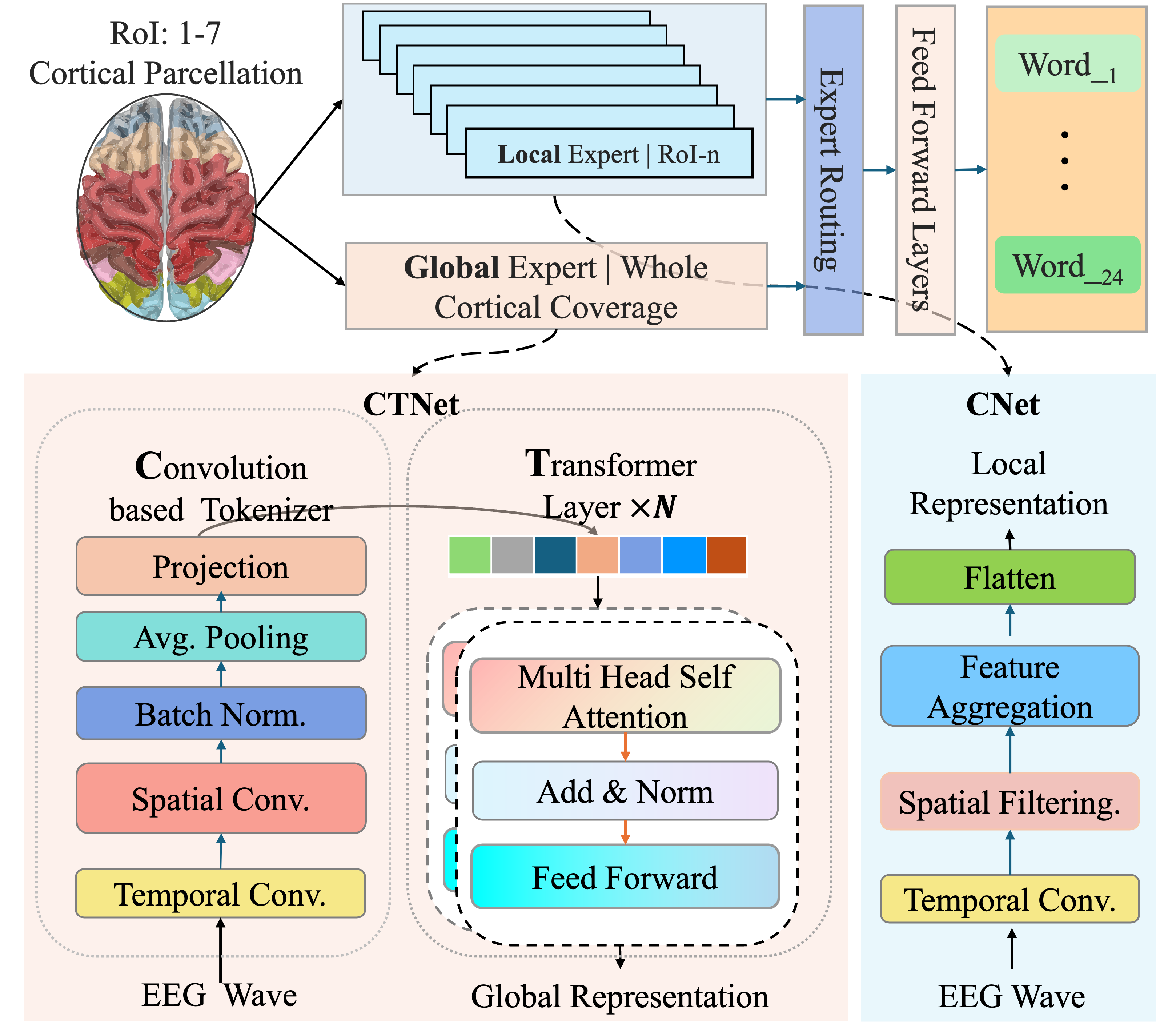}

\caption{Overview of the proposed BrainStack for EEG-based text decoding. 
Raw EEG signals \(\mathbf{X} \in \mathbb{R}^{C \times T}\) are partitioned into seven region-specific subsets and one global input. 
Local experts adopt lightweight spatio-temporal CNNs, while the global expert (CTNet) combines CNN and Transformer layers. 
Outputs are adaptively fused by an expert routing mechanism \(f_{\text{meta}}\) for final prediction \(y \in \mathcal{Y}\).}
    \label{fig:model}
    \vspace{-0.75em}
\end{figure}

\subsection{Problem Formulation}

We formulate EEG-based text decoding as a structured multi-class classification task. Each EEG trial is represented as a matrix $\mathbf{X} \in \mathbb{R}^{C \times T}$, where $C$ denotes the number of channels and $T$ is the number of time steps. The objective is to identify the corresponding cognitive or behavioral class label $y \in \{1, \ldots, K\}$. Unlike images or speech, EEG signals are inherently structured by anatomical priors. Channels can be grouped into $N$ functionally defined brain regions, each associated with specific cognitive subfunctions and distinct signal characteristics. Let $\mathbf{X}_i$ denote the input segment corresponding to region $i$, and let its task-relevant latent representation $\mathcal{I}_{\text{local}}^i$ be defined as:

\begin{equation}
\mathcal{I}_{\text{local}}^i = \phi_i(\mathbf{X}_i),
\label{eq:local_repr}
\end{equation}

where $\phi_i(\cdot)$ is a region-specific encoder. Due to anatomical and functional specialization, the regional representations are partially independent in their distribution and semantics.




Cognitive processes depend on coordinated activity across anatomically distinct brain regions, making inter-regional complementarity essential for capturing task-relevant semantics. Because no single region can fully represent complex cognitive states, integrating distributed neural features into a unified representation is necessary. To accommodate task-specific variability—such as the dominance of temporal regions in language decoding or central regions in motor imagery—we model the global neural representation as a weighted aggregation of region-specific features, formally defined as:

\begin{equation}
\mathcal{I}_y = \sum_{i=1}^{N} \omega_i^y \cdot \mathcal{I}_{\text{local}}^i = \sum_{i=1}^{N} \omega_i^y \cdot \phi_i(\mathbf{X}_i),
\label{eq:task_specific}
\end{equation}

where $\mathcal{I}_y$ represents the \textit{task-specific global representation} that integrates region-specific features according to their contextual relevance under task $y$. The term $\omega_i^y$ denotes the learnable importance weight assigned to region $i$, reflecting its contribution to the task. These weights are dynamically optimized to emphasize anatomically distinct brain regions based on their neural significance. 

Finally, the key challenge of EEG decoding lies in balancing regional specialization with cross-region integration, while adapting to task-specific variability. Formally, the decoding objective is to simultaneously learn both the local encoders $\phi_i(\cdot)$ and a fusion function $g(\cdot)$ to yield:

\begin{equation}
\hat{y} = \arg\max_k \; \Bigg[g\left(  \sum_{i=1}^{N} \omega_i^y \cdot \phi_i(\mathbf{X}_i)  \right)\Bigg]_k,
\label{eq:final_decoding}
\end{equation}
where $k$ indexes the set of class labels, and $g(\cdot)$ maps the integrated features into class logits. The $\arg \max$ operator selects the class with the highest predicted score, resulting in the final output.

\subsection{The Proposed Framework: BrainStack}



Motivated by the principle that cognition arises from the interplay between distributed global coordination and localized regional computation \cite{yue2017complexity}, BrainStack formulates EEG decoding within a Neuro–Mixture-of-Experts paradigm. Specifically, we explicitly decompose each EEG trial into a global view and seven functionally defined regional views, following canonical anatomical divisions: Prefrontal, Frontal, Central, Left-Temporal, Right-Temporal, Parietal, and Occipital \cite{liang2024transfer}. Formally, each EEG sample $\mathbf{X} \in \mathbb{R}^{C \times T}$ is processed in parallel by a global expert and a set of regional experts, where each expert receives a spatially tailored subset $\mathbf{X}_i \in \mathbb{R}^{C_i \times T}$. This modular decomposition lays the foundation for adaptive expert routing and hierarchical coordination in the subsequent stages of BrainStack. 

To ensure each regional expert specialize in extracting discriminative, anatomically grounded features while supporting scalable parallel computation across cortical regions. All experts are trained jointly in an end-to-end manner, facilitating coordinated learning and enabling BrainStack’s Neuro-MoE to effectively integrate both localized and global information~\cite{chen2024robust}.

\paragraph{CTNet: Attention-based Global Expert}
As illustrated in Fig.~\ref{fig:model}, the global expert is responsible for modeling long-range dependencies and whole-brain contextual interactions. To achieve this, CTNet combines convolutional feature extraction with Transformer-based sequence modeling, enabling it to capture both fine-grained temporal dynamics and cross-regional coordination across the full electrode montage. The input to the global expert is a preprocessed EEG trial $\mathbf{X} \in \mathbb{R}^{C \times T}$, where $C$ denotes the number of channels and $T$ the number of time steps. CTNet processes this input through two primary stages: a convolutional patch embedding module that transforms raw EEG segments into latent tokens, and a Transformer encoder that integrates information across these tokens to form a global, cognitively meaningful representation.

\textbf{1) Convolutional Patch Embedding.} 
The global expert first converts the continuous EEG waveform into a sequence of latent tokens via a convolutional patch embedding module. A temporal convolution extracts short-duration dynamics within each channel, followed by a spatial convolution that aggregates information across electrodes to capture whole-brain coordination. Batch normalization and ELU activation stabilize training, and average pooling partitions the signal into fixed-length windows, each forming a multi-channel temporal patch. A final $1 \times 1$ convolution projects each patch into the latent space, yielding a token sequence $\mathbf{Z} \in \mathbb{R}^{N \times D}$ that is subsequently processed by the Transformer encoder.

\textbf{2) Transformer Encoder.} 
The resulting patch tokens are processed by a stack of Transformer layers, each composed of multi-head self-attention and feedforward blocks with residual Add–Norm connections. This stage performs global integration across the entire token sequence, enabling the model to capture long-range temporal dependencies and cross-channel interactions that extend beyond the receptive field of convolutional filters.

Through this process, the global expert produces a set of contextualized representations that summarize whole-brain spatiotemporal dynamics. These global representations are subsequently passed to the expert routing gate, where they are jointly considered with the outputs of the regional experts to determine the final routed prediction.

\paragraph{CNet: Convolution-based Local Expert}
To complement the global expert, BrainStack incorporates a set of lightweight regional experts, each dedicated to modeling EEG signals from a specific functional cortical area. For each region $n \in {1, ..., N}$ (with $N = 7$), the corresponding input $\mathbf{X}_n \in \mathbb{R}^{C_n \times T}$ is extracted via an anatomically informed channel mapping (standard 10-10/10-20 system), ensuring that $C_n \ll C$ and enabling each expert to focus on localized neural dynamics within its region.

Inspired by the architecture proposed in ~\cite{lawhern2018eegnet}, each regional expert adopts a compact convolutional design tailored for EEG decoding. The module consists of temporal convolutions that capture short-duration temporal structure, depthwise spatial convolutions that learn inter-channel correlations within the region, and separable convolutions that provide efficient feature abstraction. Batch normalization and dropout are applied throughout to ensure stable optimization and mitigate overfitting.

\paragraph{Adaptive Expert Routing Gate}
To coordinate the global and regional experts, BrainStack employs an adaptive expert routing gate. The router assigns context-dependent routing weights to each expert and aggregates their representations accordingly:
\begin{equation}
F_{\text{meta}} = \sum_{i=1}^N \alpha_i \cdot F_i,
\end{equation}
where $F_i$ denotes the representation produced by the $i$-th expert and $\alpha_i$ is the routing coefficient produced by the gate. The coefficients are normalized via a softmax constraint, ensuring $\sum_{i=1}^N \alpha_i = 1$. 

Each routing weight $\alpha_i$ is computed through a lightweight scoring function:
\begin{equation}
\alpha_i = \frac{\exp(h(F_i))}{\sum_{j=1}^{N} \exp(h(F_j))},
\end{equation}
where $h(\cdot)$ is a learnable projection that evaluates the relevance of each expert’s representation to the final decision. This mechanism allows the Neuro-MoE to dynamically shift focus among experts, effectively highlighting spatially or temporally informative regions while attenuating noisy or uninformative ones. The routed representation $F_{\text{route}}$ is then passed to a final prediction head, implemented as a fully connected layer followed by softmax activation.

\begin{table*}[t]
\centering
\caption{Comparison of SS-EEG with representative EEG datasets for imagined or silent speech decoding.}
\label{tab:dataset_comparison}
\renewcommand{\arraystretch}{1.2}
\resizebox{\textwidth}{!}{%
\begin{tabular}{lccccccc}
\toprule
\textbf{Dataset} & \textbf{Subjects} & \textbf{Vocabulary / Classes} & \textbf{Trials} & \textbf{Duration} & \textbf{Channels} & \textbf{Sessions} & \textbf{Repeats per Class} \\
\midrule
KaraOne~\cite{zhao2015classifying} & 12 (8 usable) & 11 (7 phonemes + 4 words) & 1056 & $\sim$4.5 h & 64 EEG + 4 EOG & 3 & 12 \\
Thinking Out Loud~\cite{nieto2022thinking} & 10 & 4 (Spanish words) & 2236 & $\sim$3.5 h & 128 EEG + 8 EXG\textsuperscript{$\dagger$} & 3 & 55--60 \\
\textbf{SS-EEG (Ours)} & 12 (10 usable) & 24 (English words) & 60{,}000 & $\sim$120 h & 128 EEG + 8 EXG\textsuperscript{$\dagger$} & 16 & 250 \\
\bottomrule
\end{tabular}
}
\begin{tablenotes}
\footnotesize
\item[$\dagger$] EXG channels include 4 auxiliary electrodes for eye (EOG) and 2 muscle (EMG) activity monitoring, as well as 2 reference electrodes.
\end{tablenotes}
\end{table*}

\subsection{Hierarchical Multi-Objective Optimization}

BrainStack employs a hierarchical multi-objective training strategy that aligns all experts within a unified Neuro-MoE framework. Training is scheduled dynamically, beginning with a global warm-up phase to establish a stable whole-brain representation, and gradually transitioning into full multi-expert supervision encompassing fused prediction, global modeling, regional specialization, and cross-regional distillation~\cite{du2020agree, zoumpourlis2022motor}. 
Formally:
\begin{equation}
\mathcal{L}_{\text{total}} = \lambda \cdot \mathcal{L}_{\text{fused}} + \alpha \cdot \mathcal{L}_{\text{global}} + \beta \cdot \mathcal{L}_{\text{local}} + \gamma \cdot \mathcal{L}_{\text{distill}},
\label{eq:total_loss}
\end{equation}
where $\lambda$, $\alpha$, $\beta$, and $\gamma$ are hyperparameters controlling the relative weights of each term.

We introduce a dynamic scheduling function to progressively adjust task weights during training. Let $\mathcal{P}(t)$ denote the training progress at epoch $t$:

\begin{equation}
\mathcal{P}(t) = \min \left(1.0, \,\frac{\text{epoch} - T_{\text{warmup}}}{T_{\text{transition}}} \right),
\end{equation}
where $T_{\text{warmup}}$ is the duration of the warm-up stage and $T_{\text{transition}}$ controls the duration of the transition phase.

The fused loss weight is then annealed according to:

\begin{equation}
\lambda = (1 - \mathcal{P}(t)) \cdot \lambda_{\text{min}} + \mathcal{P}(t) \cdot \lambda_{\text{max}},
\end{equation}

To regulate auxiliary components, we define a shared scheduling function:
\begin{equation}
\begin{aligned}
\omega_{\text{aux}}(x) 
&= x \cdot \mathcal{P}(t) \\
&\quad \cdot \Biggl(1 - \min\Biggl(
\frac{\mathcal{L}_{\text{fused}}}{\text{max\_loss\_estimate}}, \, 1.0
\Biggr)\Biggr)
\end{aligned}
\end{equation}
where $\omega_{\text{aux}}(x)$ denotes the dynamic weight of auxiliary supervision and $\text{max\_loss\_estimate}$ is an empirical upper bound on the loss. 

The final auxiliary weights are thus:
\begin{equation}
\mathbf{w} = \big[ \omega_{\text{aux}}(\mathbf{\alpha}_{\max}), \, \space \omega_{\text{aux}}(\mathbf{\beta}_{\max}), \, \omega_{\text{aux}}(\mathbf{\gamma}_{\max}) \cdot \mathcal{P}(t)\big],
\end{equation}

\paragraph{Hierarchical Distillation Across Experts.}
To promote coordinated learning within the Neuro-MoE, we adopt a top-down distillation strategy in which the global expert provides semantic guidance to each regional expert. The distillation loss is defined as:

\begin{equation}
\begin{aligned}
\mathcal{L}_{\text{distill}} 
&= \sum_{i=1}^{N} \text{KL}\Biggl(
\text{Softmax}\!\left(\tfrac{\mathbf{F}_{\text{global}}}{T}\right) \\
&\quad\parallel\; 
\text{Softmax}\!\left(\tfrac{\mathbf{F}_i}{T}\right)
\Biggr)
\end{aligned}
\end{equation}

where $\mathbf{F}_{\text{global}}$ and $\mathbf{F}_i$ are the logits from the global and $i$-th regional expert, respectively, and $T$ is a temperature parameter.
This hierarchical objective transfers high-level semantic structure from the global expert to each regional expert, improving their capacity to model fine-grained local patterns while preserving coherence with the fused routed prediction.

\begin{table*}[t]
  \caption{Performance comparison on the SS-EEG dataset for word-level classification.}
  \label{tab:main_results}
  \centering
  \renewcommand{\arraystretch}{1.3}
  \resizebox{\textwidth}{!}{%
  \begin{tabular}{lcccccccccccc}
    \toprule
    Model & Param. & S01 & S02 & S03 & S04 & S05 & S06 & S07 & S08 & S09 & S10 & Avg. Acc \\
    \midrule
    EEGNet            & \textbf{8.54K}  & 40.80\% & 35.30\% & 33.50\% & 22.90\% & \textbf{17.70\%} & 10.60\% & 49.60\% & 14.00\% & 38.70\% & 24.70\% & 28.78\% \\
    TCNet             & 78.62K          & 36.90\% & 36.40\% & 32.70\% & 22.70\% & 14.30\% & 15.30\% & 44.40\% & 16.40\% & 50.40\% & 25.50\% & 29.50\% \\
    EEGConformer      & 0.75M           & 58.20\% & 32.70\% & 27.50\% & 12.20\% & 11.20\% & 7.00\%  & 30.90\% & 7.00\%  & 23.90\% & 28.30\% & 23.89\% \\
    STTransformer     & 2.78M           & 60.30\% & 16.40\% & 30.90\% & 21.30\% & 17.40\% & \textbf{35.10\%} & 15.60\% & \textbf{31.20\%} & 31.20\% & 23.40\% & 28.28\% \\
    LaBraM            & 5.80M           & 63.40\% & 13.00\% & 15.10\% & 9.10\%  & 8.60\%  & 7.50\%  & 34.30\% & 8.30\%  & 13.20\% & 10.40\% & 18.29\% \\
    \cdashline{1-13}[1pt/2pt]
    BrainStack\_Homo  & 5.93M           & 81.04\% & 27.79\% & 32.98\% & 23.37\% & 6.23\%  & 11.69\% & 53.77\% & 11.69\% & 60.77\% & 18.44\% & 32.78\% \\
    BrainStack\_RoI5  & 1.05M           & 80.26\% & 39.22\% & 44.15\% & 24.94\% & 5.71\%  & 11.42\% & 56.10\% & 15.84\% & 62.85\% & 31.42\% & 37.19\% \\
    \textbf{BrainStack} & 1.06M        & \textbf{88.05\%} & \textbf{39.48\%} & \textbf{48.57\%} & \textbf{29.87\%} & 10.13\% & 17.40\% & \textbf{60.25\%} & 19.74\% & \textbf{70.64\%} & \textbf{34.54\%} & \textbf{41.87\%} \\
    \bottomrule
  \end{tabular}}
\end{table*}

\section{Experiments}
\subsection{Dataset}
\label{subsection: experiment}

\paragraph{Silent Speech EEG (SS-EEG)} To evaluate the proposed Neuro-MoE framework, we introduce SS-EEG, a large-scale dataset tailored for word-level silent speech decoding from EEG signals. SS-EEG contains over 120 hours of recordings from 12 participants, each performing covert articulation of 24 commonly used English words (e.g., “go”, “happy”, “python”) across 16 sessions. With approximately 6,000 trials per subject, the dataset offers dense temporal coverage and robust within- and cross-subject evaluation settings. To the best of our knowledge, SS-EEG is the largest publicly available corpus for word-level silent speech decoding using EEG, and will be released upon publication.

A single session contains 375 silent speech trials and lasts approximately 40 minutes including scheduled breaks. The experimental paradigm follows a four-stage protocol: \emph{rest}, \emph{read}, \emph{preparation}, and \emph{silent speech}. Each trial begins with a 5-second rest phase, including a 1.5-second fixation cross. The target word is then displayed for 1 second during the read phase. A brief 0.2-second auditory cue signals transition into the imagination phase, where the subject silently repeats the word for 1.5 seconds. To minimize fatigue, participants are provided with a break after every 20 trials for up to 5 minutes. EEG signals were recorded using a 128-channel Neuroscan system at a 1000 Hz sampling rate. Signals were band-pass filtered to 1–45 Hz and segmented into clean trial epochs. We further applied per-channel z-score normalization and removed trials with excessive noise or artifacts. It is noted that due to severe signal contamination in two subjects, the final released dataset includes recordings from 10 participants. For a comprehensive description of the data acquisition protocol and preprocessing pipeline, we refer readers to Section “The SS-EEG Dataset” in the supplementary material.

\paragraph{Thinking out Loud} Beyond the primary within-subject evaluation on SS-EEG, we also conduct experiments under a cross-subject protocol as well as on the public Thinking Out Loud dataset~\cite{nieto2022thinking}. These results, reported in the Appendix under Additional Experiments: Cross-Subject and Public Dataset, further validate the robustness and generalization capacity of the proposed Neuro-MoE framework.

\begin{figure*}[t]
    \centering
    \includegraphics[width=0.9\textwidth]{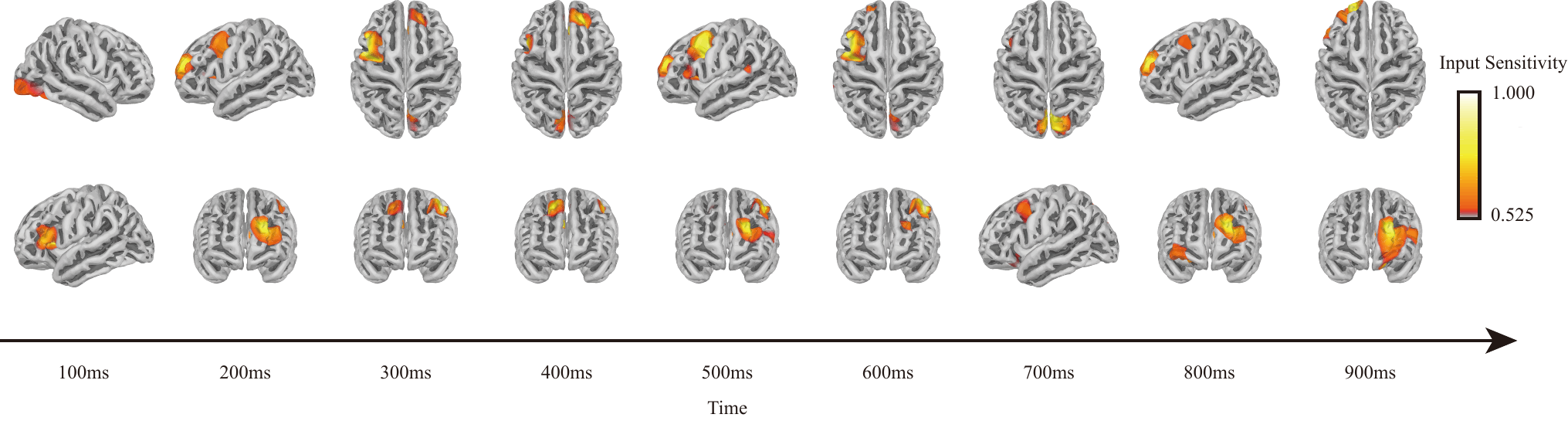}
    \vspace{-0.75em}
    \caption{\textbf{Human attention dynamics during silent speech execution.} 
    Temporal evolution of brain activation maps obtained through attribution analysis. 
    The maps illustrate the spatiotemporal distribution of sensitivity in response to the input stimulus across different brain regions from 100ms to 900ms. 
    Warmer colors represent higher sensitivity, indicating stronger contribution to the model's decision at each time step. 
    Each vertical pair of cortical maps represents two complementary views of the brain at the same moment, offering a clearer visualization of region-specific activations. 
    The analysis captures the progression of localized neural responses transitioning into broader inter-regional integration, revealing both independent regional dynamics and global coordination underlying silent speech decoding.}
    \label{fig:heatmap}     
\end{figure*}

\begin{figure*}[ht]
    \centering
    \includegraphics[width=0.9\linewidth]{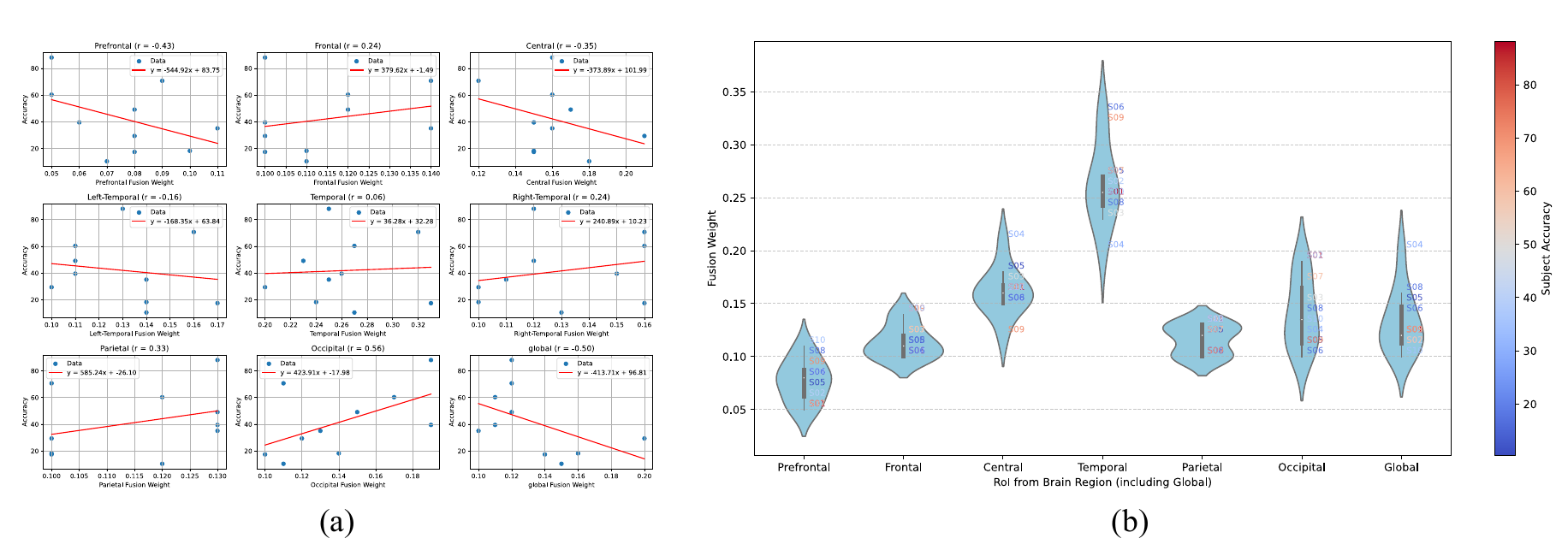}
    \vspace{-1em}
\caption{Regional contribution analysis in the adaptive routing gate. 
\textbf{(a)} Correlation between average fusion weight of each region and subject-level accuracy, with fitted regression lines and Pearson’s $r$. 
\textbf{(b)} Violin plots of weight distributions across experts, overlaid with subject accuracies (color-coded), highlighting variability in regional importance.}
    \label{fig:branch-weight-analysis}
\end{figure*}

\subsection{Evaluation Metric}
\label{subsec:evaluation_metric}
\paragraph{Evaluation metric.}
We adopt a session-wise evaluation protocol to avoid data leakage. For each subject, the 16 EEG sessions are split into 14 for training, 1 for validation, and 1 for testing, , ensuring that all splits originate from non-overlapping temporal blocks. Model performance is reported as the average classification accuracy across subjects, following the same evaluation protocol for all baselines.

\paragraph{Baselines.} We benchmark BrainStack against a representative suite of state-of-the-art EEG decoding models. The comparison includes two fully supervised CNN-based architectures—EEGNet~\cite{lawhern2018eegnet} and TCNet~\cite{ingolfsson2020eeg}—two Transformer-based models—EEGConformer~\cite{song2022eeg} and STTransformer~\cite{song2021transformer}—and the pre-trained foundation model LaBraM~\cite{jiang2024large}.

\paragraph{Implementation Details.}

The model is implemented in PyTorch and trained on one NVIDIA A40 GPU using mixed precision. We use stochastic gradient descent (SGD) optimizer with a learning rate of $5 \times 10^{-3}$, momentum of $0.9$ and weight decay of $1 \times 10^{-4}$. The training objective is defined as a dynamic multi-objective loss~\eqref{eq:total_loss}, combining cross-entropy for classification and KL divergence for distillation with a temperature of 4.0~\cite{tu2022dynamic, hamidi2024train}. Following a 5-epoch warm-up, the fusion loss weight increases linearly from 0.2 to 1.0, while the global supervision weight decays from 0.8 to 0. Regional and distillation weights are further modulated by training progress and the current fused loss. Training proceeds for up to 100 epochs, controlled by early stopping with the patience of 5 epochs.

\subsection{Performance Comparison}
\label{subsec:performance_comparison}
Table~\ref{tab:main_results} summarizes the classification performance of BrainStack and a diverse set of baseline models on the SS-EEG dataset. Among the baselines, EEGNet and TCNet represent compact CNN-based architectures, while EEGConformer and STTransformer are attention-based designs. LaBraM serves as a large-scale pretrained foundation model.

BrainStack achieves the best overall performance with an average accuracy of 41.87\%, outperforming all baselines and model variants. Compared to the strongest baseline (TCNet, 29.50\%), BrainStack delivers a significant improvement of +12.37\%. Notably, it achieves the highest accuracy on 8 out of 10 subjects, and remains highly competitive on the remaining ones, reflecting strong generalization across individuals.

Comparisons with model variants further highlight the benefits of the proposed Neuro-MoE formulation. BrainStack\_ROI5 (37.19\%), which adopts a coarser 5-region anatomical partition, lags behind by 4.68\%, underscoring the importance of fine-grained, functionally guided cortical decomposition. Likewise, BrainStack\_Homo (32.78\%), which replaces heterogeneous experts with identical CTNet encoders across all branches, performs significantly worse despite having more than five times the parameter count. Finally, BrainStack surpasses LaBraM (18.29\%) by a wide margin while using only one-fifth of the parameters (1.06M vs.\ 5.80M), emphasizing that task-specific architectural design and hierarchical expert coordination are more effective than model size alone for low-SNR neural decoding tasks.

\begin{figure}[htbp]
    \centering
    \includegraphics[width=\linewidth]{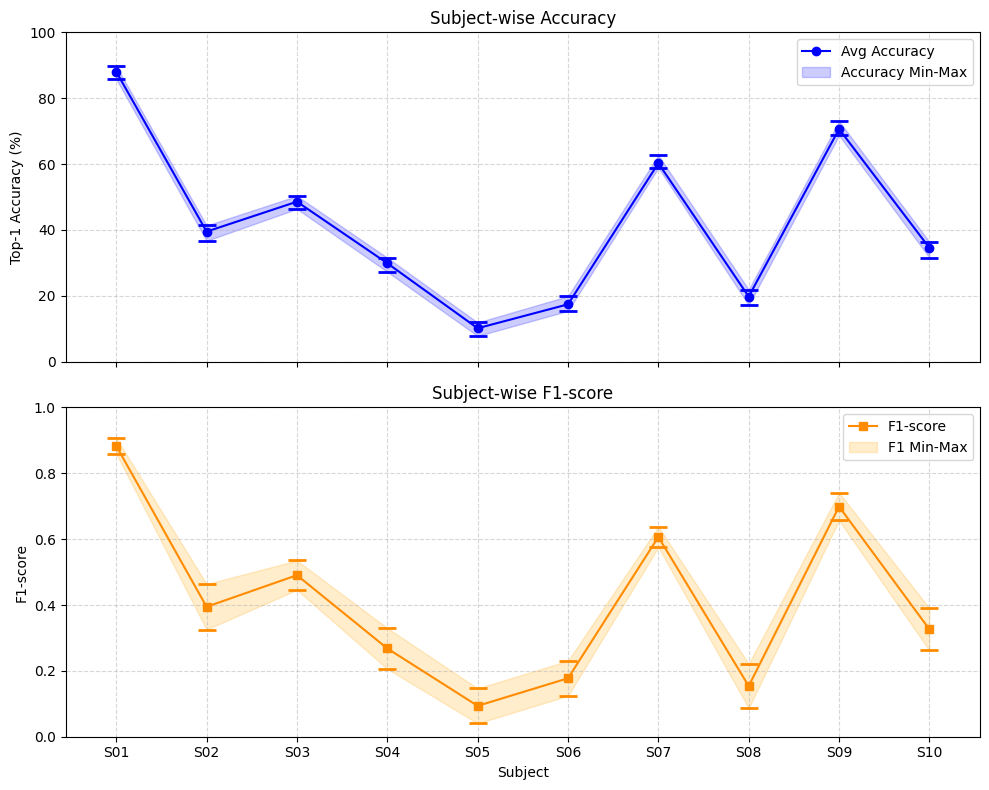}
    \caption{Subject-wise performance variability across three random seeds. 
    Top: classification accuracy with min–max error bands.
	Bottom: F1-score with corresponding min–max range.}
    \label{fig:subject_acauracy_f1}
\end{figure}

To assess subject-specific consistency, we report accuracy and F1-score for each of the 10 test subjects, averaged over three random seeds. As shown in Figure~\ref{fig:subject_acauracy_f1}, substantial inter-subject variability is observed. Overall, the error bars illustrate a clear pattern: strong stability in high-performing subjects and pronounced variability in weaker ones.  High-performing subjects (e.g., S01, S07, S09) achieve accuracies above 60\% with minimal variance, indicating stable and reliable neural patterns. Notably, S01 reaches 88.05\% accuracy and 0.88 F1, the highest across all subjects. In contrast, low-performing subjects (e.g., S05, S06, S08) exhibit both low accuracy (S05: 10.13\%) and large seed-dependent fluctuations, reflecting sensitivity to noise and inconsistent neural responses. The trend highlights the value of BrainStack’s modular expert design and adaptive routing, which help manage subject-specific neural heterogeneity.

To evaluate the contribution of each cortical region, we analyze the learned expert routing weights. As shown in Fig.~\ref{fig:branch-weight-analysis}(a), the Occipital expert exhibits the strongest positive correlation with subject-level accuracy ($r = 0.56$), highlighting the importance of visual-cortical activity for silent-speech decoding. In contrast, the Global expert shows a moderate negative correlation ($r = -0.50$), suggesting that excessive reliance on holistic features may be suboptimal. Prefrontal and Central experts show mild negative trends, whereas Parietal and Frontal experts are weakly positive. Temporal experts, though weakly correlated ($r = 0.06$), maintain consistently high routing weights across subjects. Fig.~\ref{fig:branch-weight-analysis}(b) further shows that high-performing subjects (e.g., S05, S09) assign greater weight to Temporal and Occipital experts, indicating their complementary predictive value. Overall, these patterns demonstrate that the routing gate not only prioritizes highly informative regions but also leverages stable contributors to enhance generalization.


\subsection{Ablation Study}

\begin{table*}[t]
  \caption{Ablation study on expert architecture, routing mechanism, and loss design.}
  \label{tab:ablation}
  \centering
  \renewcommand{\arraystretch}{1.5}
  \resizebox{\textwidth}{!}{%
  \begin{tabular}{llccccccccccccc}
    \toprule
    Objective & Model & Param. & S01 & S02 & S03 & S04 & S05 & S06 & S07 & S08 & S09 & S10 & Avg. Acc \\
    \midrule
    \multirow{4}{*}{\begin{tabular}[c]{@{}l@{}}Expert\\ architecture\end{tabular}} 
      & Local only-CNet            & \textbf{0.18M} & 78.44\% & 28.83\% & 22.59\% & 17.14\% & 6.23\% & 4.41\% & 53.77\% & 12.47\% & 57.14\% & 27.53\% & 30.86\% \\
      & Homogeneous-CNet           & 0.20M          & 70.65\% & 34.02\% & 36.36\% & 22.86\% & 4.94\% & 3.38\% & 55.06\% & 12.47\% & 62.08\% & 34.29\% & 33.61\% \\
      & Global only-CTNet          & 1.01M          & 72.47\% & 28.05\% & 44.41\% & 24.94\% & 8.31\% & 10.38\% & 56.62\% & 15.32\% & 65.71\% & 28.83\% & 35.50\% \\
      & Homogeneous-CTNet          & 5.93M          & 81.04\% & 27.79\% & 32.98\% & 26.75\% & 6.23\% & 11.69\% & 53.77\% & 11.69\% & 60.77\% & 18.44\% & 33.12\% \\
    \cdashline{1-14}[1pt/2pt]
    \multirow{3}{*}{\begin{tabular}[c]{@{}l@{}}Routing Mechanism \\ and Loss Design\end{tabular}} 
      & Token-level fusion         & 5.21M & 82.33\% & 29.61\% & 32.98\% & 14.28\% & 9.09\% & 11.17\% & 34.55\% & 12.73\% & 42.86\% & 24.45\% & 29.41\% \\
      & w/o knowledge distillation & 1.06M & 84.94\% & 29.87\% & 38.70\% & 17.14\% & 5.97\% & 14.03\% & 54.03\% & 14.55\% & 70.38\% & 27.27\% & 35.69\% \\
      & w/o warm-up                & 1.06M & 84.42\% & 34.54\% & 29.87\% & 12.98\% & 7.01\% & 12.47\% & 56.62\% & 16.6\% & 55.58\% & 29.61\% & 33.97\% \\
    \cdashline{1-14}[1pt/2pt]

    Ours & \textbf{BrainStack} & 1.06M & \textbf{88.05\%} & \textbf{39.48\%} & \textbf{48.57\%} & \textbf{29.87\%} & \textbf{10.13\%} & \textbf{17.40\%} & \textbf{60.25\%} & \textbf{19.74\%} & \textbf{70.64\%} & \textbf{34.54\%} & \textbf{41.87\%} \\
    \bottomrule
  \end{tabular}}
\end{table*}

To evaluate the contributions of individual components in our framework, we conduct ablation studies on three key aspects of BrainStack:
(1) the expert architecture,
(2) the expert routing mechanism, and
(3) the hierarchical loss design.
Results across ten subjects are reported in Table~\ref{tab:ablation}.

\paragraph{Ensemble architecture.}
We examine four architectural variants to assess the impact of structural design.
The \emph{Local only–CNet} configuration aggregates predictions from independent regional CNN experts and achieves 30.86\%, confirming the limited capability of purely local modeling without global coordination. Introducing structural consistency through \emph{Homogeneous–CNet} yields a small improvement (33.61\%), while \emph{Global only–CTNet} reaches a stronger 35.50\%, validating the necessity of whole-brain contextual information. The strongest homogeneous variant, \emph{Homogeneous–CTNet} (33.12\%), applies identical CTNet encoders across all experts but still falls short of capturing distributed neural dynamics effectively. In contrast, full BrainStack attains 41.87\% by combining heterogeneous global and regional experts and aligning them through anatomically informed functional decomposition. These results demonstrate that structural diversity, expert specialization, and functionally guided modularization are more effective than either global or local modeling alone.

\paragraph{Expert routing mechanism and hierarchical loss design.}
To assess the role of routing and training objectives, we evaluate token-level fusion and partial loss configurations. The token-level fusion baseline employs an attention-based aggregator over intermediate tokens but performs poorly (29.4\%), indicating that shallow token integration is insufficient without heterogeneous experts and richer contextual modeling.

Ablations on the loss design show that removing global-to-local distillation reduces performance to 35.69\%, while omitting the warm-up schedule further drops accuracy to 33.97\%. These results highlight that adaptive expert routing, hierarchical distillation, and progressive optimization are essential for stable training and effective coordination within the Neuro-MoE.

\section{Conclusion}
We present BrainStack, the first functionally guided Neuro–Mixture-of-Experts framework for EEG-based neural language decoding. BrainStack unifies global and region-specific neural representations through heterogeneous experts and an adaptive expert routing gate, enabling anatomically grounded and context-aware integration of distributed cortical signals. To support coordinated expert learning, we introduce a hierarchical multi-objective training strategy with progressive scheduling, effectively combining global supervision, regional specialization, and cross-regional distillation. We further release SilentSpeech-EEG (SS-EEG), a large-scale benchmark containing over 120 hours of recordings, making it the most extensive dataset available for word-level silent speech decoding. Comprehensive evaluations—including within-subject, cross-subject, and additional tests on the Thinking Out Loud dataset—demonstrate that BrainStack achieves state-of-the-art performance and strong generalization.

{
    \small
    \bibliographystyle{ieeenat_fullname}
    \bibliography{main}

@String(ICASSP=	{ICASSP})

@String(AAAI = {AAAI})

@article{jiang2024neurolm,
  title={NeuroLM: A universal multi-task foundation model for bridging the gap between language and EEG signals},
  author={Jiang, Wei-Bang and Wang, Yansen and Lu, Bao-Liang and Li, Dongsheng},
  journal={arXiv preprint arXiv:2409.00101},
  year={2024}
}

@article{wang2024cbramod,
  title={Cbramod: A criss-cross brain foundation model for eeg decoding},
  author={Wang, Jiquan and Zhao, Sha and Luo, Zhiling and Zhou, Yangxuan and Jiang, Haiteng and Li, Shijian and Li, Tao and Pan, Gang},
  journal={arXiv preprint arXiv:2412.07236},
  year={2024}
}

@article{wang2024eegpt,
  title={Eegpt: Pretrained transformer for universal and reliable representation of eeg signals},
  author={Wang, Guangyu and Liu, Wenchao and He, Yuhong and Xu, Cong and Ma, Lin and Li, Haifeng},
  journal={Advances in Neural Information Processing Systems},
  volume={37},
  pages={39249--39280},
  year={2024}
}

@article{nieto2022thinking,
  title={Thinking out loud, an open-access EEG-based BCI dataset for inner speech recognition},
  author={Nieto, Nicol{\'a}s and Peterson, Victoria and Rufiner, Hugo Leonardo and Kamienkowski, Juan Esteban and Spies, Ruben},
  journal={Scientific data},
  volume={9},
  number={1},
  pages={52},
  year={2022},
  publisher={Nature Publishing Group UK London}
}

@inproceedings{zhao2015classifying,
  title={Classifying phonological categories in imagined and articulated speech},
  author={Zhao, Shunan and Rudzicz, Frank},
  booktitle={2015 IEEE international conference on acoustics, speech and signal processing (ICASSP)},
  pages={992--996},
  year={2015},
  organization={IEEE}
}

@article{ding2023lggnet,
  title={LGGNet: Learning from local-global-graph representations for brain--computer interface},
  author={Ding, Yi and Robinson, Neethu and Tong, Chengxuan and Zeng, Qiuhao and Guan, Cuntai},
  journal={IEEE Transactions on Neural Networks and Learning Systems},
  volume={35},
  number={7},
  pages={9773--9786},
  year={2023},
  publisher={IEEE}
}

@inproceedings{duan2023,
  author       = {Duan, T. and Wang, Z. and Doretto, G. and Li, F. and Tao, C. and Adjeroh, D.},
  title        = {Replay with stochastic neural transformation for online continual EEG classification},
  booktitle    = {Proceedings of the IEEE International Conference on Bioinformatics and Biomedicine (BIBM)},
  pages        = {1874--1879},
  year         = {2023}
}

@article{he2013brain,
  title={Brain-Computer Interface: Challenges and Opportunities},
  author={He, B.},
  journal={Transactions of Japanese Society for Medical and Biological Engineering},
  volume={51},
  number={Supplement},
  pages={M--148},
  year={2013},
  publisher={Japanese Society for Medical and Biological Engineering}
}

@article{liang2024,
  author       = {Liang, Z. and Zheng, Z. and Chen, W. and Pei, Z. and Wang, J. and Chen, J.},
  title        = {A novel deep transfer learning framework integrating general and domain-specific features for EEG-based brain--computer interface},
  journal      = {Biomedical Signal Processing and Control},
  volume       = {95},
  pages        = {106311},
  year         = {2024}
}

@inproceedings{duan2023replay,
  author = {Duan, T. and Wang, Z. and Doretto, G. and Li, F. and Tao, C. and Adjeroh, D.},
  title = {Replay with stochastic neural transformation for online continual EEG classification},
  booktitle = {IEEE International Conference on Bioinformatics and Biomedicine (BIBM)},
  pages = {1874--1879},
  year = {2023}
}

@article{cao2022building,
  title={Building EEG-based CAD object selection intention discrimination model using convolutional neural network (CNN)},
  author={Cao, B. and Niu, H. and Hao, J. and Wang, G.},
  journal={Advanced Engineering Informatics},
  volume={52},
  pages={101548},
  year={2022},
  publisher={Elsevier}
}

@inproceedings{zhu2023eeg2vec,
  author = {Zhu, Q. and Zhao, X. and Zhang, J. and Gu, Y. and Weng, C. and Hu, Y.},
  title = {EEG2vec: Self-supervised electroencephalographic representation learning},
  booktitle = {IEEE ICASSP},
  year = {2023}
}

@article{tang2024triplet,
  author = {Tang, Y. and Huang, W. and Liu, R. and Yu, Y.},
  title = {Learning Interpretable Brain Functional Connectivity via Self-Supervised Triplet Network With Depth-Wise Attention},
  journal = {IEEE Journal of Biomedical and Health Informatics},
  year = {2024},
  doi = {10.1109/JBHI.2024.3429169}
}

@article{svanera2021selfsupervised,
  author = {Svanera, M. and Morgan, A. and Petro, L. S. and Muckli, L.},
  title = {A self-supervised deep neural network for image completion resembles early visual cortex fMRI activity patterns for occluded scenes},
  journal = {Journal of Vision},
  volume = {21},
  number = {7},
  pages = {5},
  year = {2021},
  doi = {10.1167/jov.21.7.5}
}

@inproceedings{wu2022lowlight,
  author = {Wu, W. and Wang, W. and Jiang, K. and Xu, X. and Hu, R.},
  title = {Self-Supervised Learning on A Lightweight Low-Light Image Enhancement Model with Curve Refinement},
  booktitle = {IEEE ICASSP},
  pages = {1890--1894},
  year = {2022},
  doi = {10.1109/icassp43922.2022.9746348}
}

@article{liang2024transfer,
  author = {Liang, Z. and Zheng, Z. and Chen, W. and Pei, Z. and Wang, J. and Chen, J.},
  title = {A novel deep transfer learning framework integrating general and domain-specific features for EEG-based brain–computer interface},
  journal = {Biomedical Signal Processing and Control},
  volume = {95},
  pages = {106311},
  year = {2024},
  doi = {10.1016/j.bspc.2024.106311}
}

@misc{chen2024robust,
  author = {Chen, X. and Jia, T. and Wu, D.},
  title = {Accurate, robust and privacy-preserving brain-computer interface decoding},
  howpublished = {arXiv preprint arXiv:2412.11390},
  year = {2024}
}

@misc{li2020unsupervised,
  author = {Li, H. and Fan, Y.},
  title = {Unsupervised deep learning for individualized brain functional network identification},
  howpublished = {arXiv preprint arXiv:2012.06494},
  year = {2020}
}

@article{deng2023seizure,
  author = {Deng, Z. and Li, C. and Song, R. and Liu, X. and Qian, R. and Chen, X.},
  title = {EEG-based seizure prediction via hybrid vision transformer and data uncertainty learning},
  journal = {Engineering Applications of Artificial Intelligence},
  volume = {123},
  pages = {106401},
  year = {2023},
  doi = {10.1016/j.engappai.2023.106401}
}

@inproceedings{hao2022speed,
  author = {Hao, X. and Sun, B.},
  title = {Speed Imagery EEG Classification with Spatial-temporal Feature Attention Deep Neural Networks},
  booktitle = {IEEE ISCAS},
  pages = {3438--3442},
  year = {2022},
  doi = {10.1109/ISCAS48785.2022.9937802}
}

@misc{shevchenko2024comparative,
  author = {Shevchenko, O. and Yeremeieva, S. and Laschowski, B.},
  title = {Comparative Analysis of Neural Decoding Algorithms for Brain-Machine Interfaces},
  howpublished = {bioRxiv},
  year = {2024}
}

@inproceedings{bian2024ondevice,
  author = {Bian, S. and Kang, P. and Moosmann, J. and Liu, M. and Bonazzi, P. and Rosipal, R. and Magno, M.},
  title = {On-device Learning of EEGNet-based Network For Wearable Motor Imagery Brain-Computer Interface},
  booktitle = {Proceedings of ACM/IEEE Embedded Systems Week},
  year = {2024}
}

@misc{lou2024dbnet,
  author = {Lou, X. and Li, X. and Meng, H. and Hu, J. and Xu, M. and Zhao, Y. and Yang, J. and Li, Z.},
  title = {EEG-DBNet: A Dual-Branch Network for Temporal-Spectral Decoding in Motor-Imagery Brain-Computer Interfaces},
  howpublished = {arXiv preprint arXiv:2405.16090},
  year = {2024}
}

@inproceedings{lee2021eegtransformer,
  author    = {Lee, Y. E. and Lee, S. H.},
  title     = {EEG-Transformer: Self-attention from Transformer Architecture for Decoding EEG of Imagined Speech},
  booktitle = {2022 10th International Winter Conference on Brain-Computer Interface (BCI)},
  pages     = {1--4},
  year      = {2021}
}

@inproceedings{ding2025attentive,
  author    = {Ding, Y. and Lee, J. H. and Zhang, S. and Luo, T. and Guan, C.},
  title     = {Decoding Human Attentive States from Spatial-temporal EEG Patches Using Transformers},
  booktitle = {Proceedings of the AAAI Conference on Artificial Intelligence},
  year      = {2025}
}

@article{shi2023meet,
  author  = {Shi, E. and Yu, S. and Kang, Y. and Wu, J. and Zhao, L. and Zhu, D. and Lv, J. and Liu, T. and Hu, X. and Zhang, S.},
  title   = {MEET: A Multi-Band EEG Transformer for Brain States Decoding},
  journal = {IEEE Transactions on Biomedical Engineering},
  year    = {2023},
  doi     = {10.1109/TBME.2023.3339892}
}

@inproceedings{choi2025geometric,
  author    = {Choi, B. J.},
  title     = {Geometric Machine Learning on EEG Signals},
  booktitle = {ICASSP},
  year      = {2025}
}

@article{zheng2024discrete,
  title={Du-IN: Discrete units-guided mask modeling for decoding speech from Intracranial Neural signals},
  author={Zheng, H. and Wang, H. and Jiang, W. and Chen, Z. and He, L. and Lin, P. and Wei, P. and Zhao, G. and Liu, Y.},
  journal={Advances in Neural Information Processing Systems},
  volume={37},
  pages={79996--80033},
  year={2024}
}

@article{wang2020silent,
  author  = {Wang, Y. and Zhang, M. and Wu, R. and Gao, H. and Yang, M. and Luo, Z. and Li, G.},
  title   = {Silent Speech Decoding Using Spectrogram Features Based on Neuromuscular Activities},
  journal = {Brain Sciences},
  volume  = {10},
  number  = {7},
  pages   = {442},
  year    = {2020},
  doi     = {10.3390/brainsci10070442}
}

@inproceedings{rekrut2021semantic,
  author    = {Rekrut, M. and Sharma, M. and Schmitt, M. and Alexandersson, J. and Krüger, A.},
  title     = {Decoding Semantic Categories from EEG Activity in Silent Speech Imagination Tasks},
  booktitle = {2021 9th International Winter Conference on Brain-Computer Interface (BCI)},
  pages     = {1--7},
  year      = {2021}
}

@article{song2023sEMG,
  author  = {Song, R. and Zhang, X. and Chen, X. and Chen, X. and Chen, X. and Yang, S. and Yin, E.},
  title   = {Decoding silent speech from high-density surface electromyographic data using transformer},
  journal = {Biomedical Signal Processing and Control},
  volume  = {80},
  pages   = {104298},
  year    = {2023},
  doi     = {10.1016/j.bspc.2022.104298}
}

@article{fitriah2022survey,
  author  = {Fitriah, N. and Zakaria, H. and Rajab, T. L. E.},
  title   = {EEG-Based Silent Speech Interface and its Challenges: A Survey},
  journal = {International Journal of Advanced Computer Science and Applications},
  year    = {2022},
  doi     = {10.14569/ijacsa.2022.0131173}
}

@misc{lashley2006organization,
  author = {Lashley, K.},
  title = {Theories of brain organization focus on two distinct, but complementary principles: modularity and network connectivity},
  year = {2006}
}

@misc{stephan2006dcm,
  author = {Stephan, K. E. and Harrison, L. and Kiebel, S. J. and David, O. and Penny, W. and Friston, K.},
  title = {Dynamic causal models of neural system dynamics},
  year = {2006}
}

@article{li2022unconsciousness,
  author = {Li, S. and Chen, Y. and Ren, P. and Li, Z. and Zhang, J. and Liang, X.},
  title = {Highly connected and highly variable: A core brain network during resting state supports propofol-induced unconsciousness},
  journal = {Human Brain Mapping},
  volume = {44},
  number = {3},
  pages = {841--853},
  year = {2022}
}

@misc{johnston2024modular,
  author = {Johnston, W. and Fusi, S.},
  title = {Modular representations emerge in neural networks trained to perform context-dependent tasks},
  howpublished = {bioRxiv},
  year = {2024}
}

@inproceedings{abbasi2020modularity,
  author = {Abbasi, N. I. and Saint-Auret, S. and Hamano, J. and Chaudhury, A. and Bezerianos, A. and Thakor, N. and Dragomir, A.},
  title = {EEG Functional Modularity Analysis Reveals Differences in Perception of Positively-Valenced Stimuli},
  booktitle = {Springer Lecture Notes in Computer Science},
  year = {2020}
}

@article{ding2022sensory,
  author = {Ding, K. and Chen, Y. and Bose, R. and Osborn, L. E. and Dragomir, A. and Thakor, N. V.},
  title = {Sensory stimulation modulates adaptability of cortical large-scale systems},
  journal = {Scientific Reports},
  volume = {12},
  year = {2022}
}

@misc{yue2017complexity,
  author = {Yue, Q. and Martin, R. and Fischer-Baum, S. and Ramos-Nuñez, A. I. and Ye, F. and Deem, M. W.},
  title = {Brain Modularity Mediates the Relation between Task Complexity and Performance},
  howpublished = {bioRxiv},
  year = {2017}
}

@misc{duggento2022intertwined,
  author = {Duggento, A. and De Lorenzo, M. and Bargione, S. and Conti, A. and Catrambone, V. and Valenza, G. and Toschi, N.},
  title = {An intertwined neural network model for EEG classification in brain-computer interfaces},
  howpublished = {arXiv preprint arXiv:2208.08860},
  year = {2022}
}

@misc{du2020agree,
  title={Agree to disagree: Adaptive ensemble knowledge distillation in gradient space},
author = {Du, S. and You, S. and Li, X. and Wu, J. and Wang, F. and Qian, C. and Zhang, C.},
  journal = {advances in neural information processing systems},
  volume={33},
  pages={12345--12355},
  year={2020}
}

@misc{zoumpourlis2022motor,
  title={Motor imagery decoding using ensemble curriculum learning and collaborative training},
author = {Zoumpourlis, G. and Patras, I.},
  booktitle={2024 12th International Winter Conference on Brain-Computer Interface (BCI)},
  pages={1--8},
  year={2024},
  organization={IEEE}
}

@article{tu2022dynamic,
  author  = {Tu, Z. and Liu, X. and Xiao, X.},
  title   = {A General Dynamic Knowledge Distillation Method for Visual Analytics},
  journal = {IEEE Transactions on Image Processing},
  volume  = {31},
  pages   = {6517--6531},
  year    = {2022}
}

@inproceedings{hamidi2024train,
  title={How to train the teacher model for effective knowledge distillation},
  author={Hamidi, Shayan Mohajer and Deng, Xizhen and Tan, Renhao and Ye, Linfeng and Salamah, Ahmed Hussein},
  booktitle={European Conference on Computer Vision},
  pages={1--18},
  year={2024},
  organization={Springer}
}

@article{lawhern2018eegnet,
  title={EEGNet: A compact convolutional neural network for EEG-based brain--computer interfaces},
  author={Lawhern, Vernon J and Solon, Amelia J and Waytowich, Nicholas R and Gordon, Stephen M and Hung, Chou P and Lance, Brent J},
  journal={Journal of neural engineering},
  volume={15},
  number={5},
  pages={056013},
  year={2018}
}

@inproceedings{ingolfsson2020eeg,
  title={EEG-TCNet: An accurate temporal convolutional network for embedded motor-imagery brain--machine interfaces},
  author={Ingolfsson, Thorir Mar and Hersche, Michael and Wang, Xiaying and Kobayashi, Nobuaki and Cavigelli, Lukas and Benini, Luca},
  booktitle={2020 IEEE International Conference on Systems, Man, and Cybernetics (SMC)},
  pages={2958--2965},
  year={2020},
  organization={IEEE}
}

@article{song2022eeg,
  title={EEG conformer: Convolutional transformer for EEG decoding and visualization},
  author={Song, Yonghao and Zheng, Qingqing and Liu, Bingchuan and Gao, Xiaorong},
  journal={IEEE Transactions on Neural Systems and Rehabilitation Engineering},
  volume={31},
  pages={710--719},
  year={2022}
}

@article{song2021transformer,
  title={Transformer-based spatial-temporal feature learning for EEG decoding},
  author={Song, Yonghao and Jia, Xueyu and Yang, Lie and Xie, Longhan},
  journal={arXiv preprint arXiv:2106.11170},
  year={2021}
}

@article{jiang2024large,
  title={Large brain model for learning generic representations with tremendous EEG data in BCI},
  author={Jiang, Wei-Bang and Zhao, Li-Ming and Lu, Bao-Liang},
  journal={arXiv preprint arXiv:2405.18765},
  year={2024}
}

@article{jamaliSemanticEncodingLanguage2024,
  title = {Semantic Encoding during Language Comprehension at Single-Cell Resolution},
author = {Jamali, M. and Grannan, B. and Cai, J. and Khanna, A. R. and Mu{\~n}oz, W. and Caprara, I. and Paulk, A. C. and Cash, S. S. and Fedorenko, E. and Williams, Z. M.},
  year = {2024},
  month = jul,
  journal = {Nature},
  volume = {631},
  number = {8021},
  pages = {610--616},
  issn = {0028-0836, 1476-4687},
  doi = {10.1038/s41586-024-07643-2},
  urldate = {2024-09-04},
}

@article{luoStableDecodingSpeech2023,
    title={Stable decoding from a speech BCI enables control for an individual with ALS without recalibration for 3 months},
  author = {Luo, S. and Angrick, M. and Coogan, C. and Candrea, D. N. and Wyse-Sookoo, K. and Shah, S. and Rabbani, Q. and Milsap, G. W. and Weiss, A. R. and Anderson, W. S. and others},
  journal={Advanced Science},
  volume={10},
  number={35},
  pages={2304853},
  year={2023},
  publisher={Wiley Online Library}
}

@article{liuDecodingSynthesizingTonal2023,
  title = {Decoding and Synthesizing Tonal Language Speech from Brain Activity},
author = {Liu, Y. and Zhao, Z. and Xu, M. and Yu, H. and Zhu, Y. and Zhang, J. and Bu, L. and Zhang, X. and Lu, J. and Li, Y. and Ming, D. and Wu, J.},
  year = {2023},
  month = jun,
  journal = {Science Advances},
  volume = {9},
  number = {23},
  pages = {eadh0478},
  issn = {2375-2548},
  doi = {10.1126/sciadv.adh0478},
  urldate = {2023-06-15},
}

@article{anumanchipalliSpeechSynthesisNeural2019,
 title={Speech synthesis from neural decoding of spoken sentences},
  author = {Anumanchipalli, G. K. and Chartier, J. and Chang, E. F.},
  journal={Nature},
  volume={568},
  number={7753},
  pages={493--498},
  year={2019},
  publisher={Nature Publishing Group UK London}
}

@article{jiang2025neural,
  title={Neural Spelling: A Spell-Based BCI System for Language Neural Decoding},
author = {Jiang, X. and Zhou, C. and Duan, Y. and Zhao, Z. and Do, T. and Lin, C.-T.}
,
  journal={arXiv preprint arXiv:2501.17489},
  year={2025}
}

@inproceedings{cao2021brain,
  title={Brain decoding using fnirs},
  author = {Cao, L. and Huang, D. and Zhang, Y. and Jiang, X. and Chen, Y.},
  booktitle={Proceedings of the AAAI Conference on Artificial Intelligence},
  volume={35},
  number={14},
  pages={12602--12611},
  year={2021}
}
}

\end{document}